\pdfoutput=1
\documentclass[conference]{IEEEtran}
\usepackage[nocompress]{cite}
\usepackage{amsmath,amssymb,amsfonts}
\usepackage{algorithmic}
\usepackage{algorithm}
\usepackage{graphicx}
\usepackage{textcomp}
\usepackage{xcolor}
\usepackage{booktabs}
\usepackage{tabularx}
\usepackage{multirow}
\usepackage{hyperref}
\usepackage{subcaption}

\makeatletter
\def\abstract{\normalfont
    \if@twocolumn
      \@IEEEabskeysecsize\bfseries\textit{\abstractname}:\ \relax
    \else
      \bgroup\par\addvspace{0.5\baselineskip}\centering\vspace{-1.78ex}\@IEEEabskeysecsize\textbf{\abstractname}\par\addvspace{0.5\baselineskip}\egroup\quotation\@IEEEabskeysecsize
    \fi\@IEEEgobbleleadPARNLSP}
\def\IEEEkeywords{\normalfont
    \if@twocolumn
      \@IEEEabskeysecsize\bfseries\textit{\IEEEkeywordsname}:\ \relax
    \else
      \bgroup\par\addvspace{0.5\baselineskip}\centering\@IEEEabskeysecsize\textbf{\IEEEkeywordsname}\par\addvspace{0.5\baselineskip}\egroup\quotation\@IEEEabskeysecsize
    \fi\@IEEEgobbleleadPARNLSP}
\makeatother

\graphicspath{{figures/}}
\begin{document}

\title{The Inference Engineering Pareto Atlas:\\ Which Optimizations Dominate the\\ Cost, Quality, and Latency Frontier?}

\author{
\IEEEauthorblockN{Srikanta Datta Tumkur, Jay Iyer, Mehar Simhadri,\\ Sai Pavan Kumar, Sai Kapil Kumar, Ramesh Nampelly}
\IEEEauthorblockA{Vizuara}
}

\maketitle

\begin{abstract}
Every optimization for LLM inference arrives with its own speedup number, and the numbers do not compare. Quantization papers report one gain, KV-cache compression papers another, and speculative decoding, batching, and sparse attention each report their own, nearly always on different models, GPUs, prompts, and quality metrics. A practitioner with a fixed budget, quality floor, or latency target cannot add these numbers up, and cannot tell which combination wins once the methods are stacked. This paper builds a cost-quality-latency Pareto atlas: every optimization and every promising combination on common axes, with the dominant configuration named in each deployment regime. The configuration space is too large to measure exhaustively, so we measure anchors on real hardware and calibrate a profiled simulator to fill the rest. The anchors are 54 configurations of Qwen2.5-7B-Instruct served by vLLM 0.12 on L4, A100, and H100 instances, 18 per GPU across three campaigns; the calibrated simulator reproduces them at the anchored batch sizes, and cross-campaign drift is below 1.5\%. A separate quality arm scores FP16, AWQ-4bit, FP8 weights, and an FP8 KV cache on GSM8K (200 questions, 5-shot). Sparse attention is the one method we could only simulate. On the calibrated grid, 18 of 36 configurations reach the frontier, and stacked combinations reach it more often than single methods (9 of 15 against 9 of 21). Measuring quality changes who wins. AWQ-4bit buys the most speed, down to 0.34$\times$ per-token latency on the L4, but loses 5.9\% of strict GSM8K accuracy and misses the 95\% quality floor by a margin inside the sampling error; the loss is in answer formatting rather than arithmetic, since flexible extraction ties the FP16 baseline. FP8 weight quantization keeps 99.4\% of baseline accuracy at 0.61 to 0.65$\times$ latency on all three GPUs and is part of three of the four regime winners. A naive FP8 KV cache posts normal throughput and answers 0 of 200 questions correctly, a configuration a speed-only benchmark would recommend and the atlas eliminates. Under two prompt designs, n-gram speculative decoding measures at 0.90 to 0.98$\times$ and adds nothing on this stack. The winner moves with the constraint and the GPU: the H100 takes the tight-latency regime, and the A100 takes throughput and low cost at \$0.106 per million tokens.
\end{abstract}


\section{Introduction}

\begin{figure*}[t]
\centering
\includegraphics[width=0.98\textwidth]{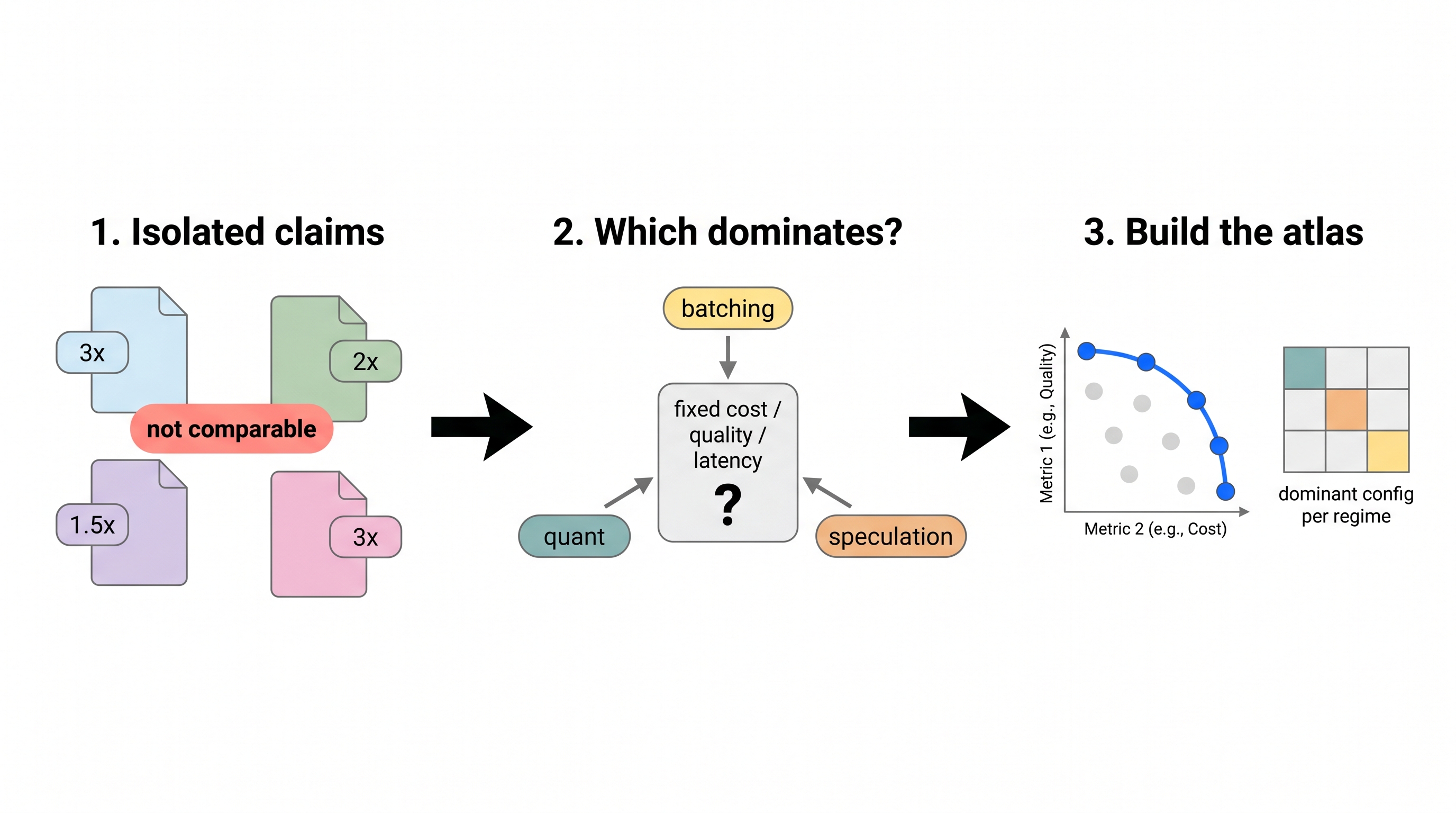}
\caption{Study overview. (1)~The field reports many isolated speedups under incompatible settings, so they cannot be compared or combined. (2)~A practitioner under a fixed cost, quality, or latency constraint needs to know which optimization or combination dominates. (3)~We place every method and promising combination on common cost-quality-latency axes, compute the Pareto frontier, and build an atlas of which operating points dominate in each regime.}
\label{fig:overview}
\end{figure*}

\begin{figure}[t]
\centering
\includegraphics[width=\columnwidth]{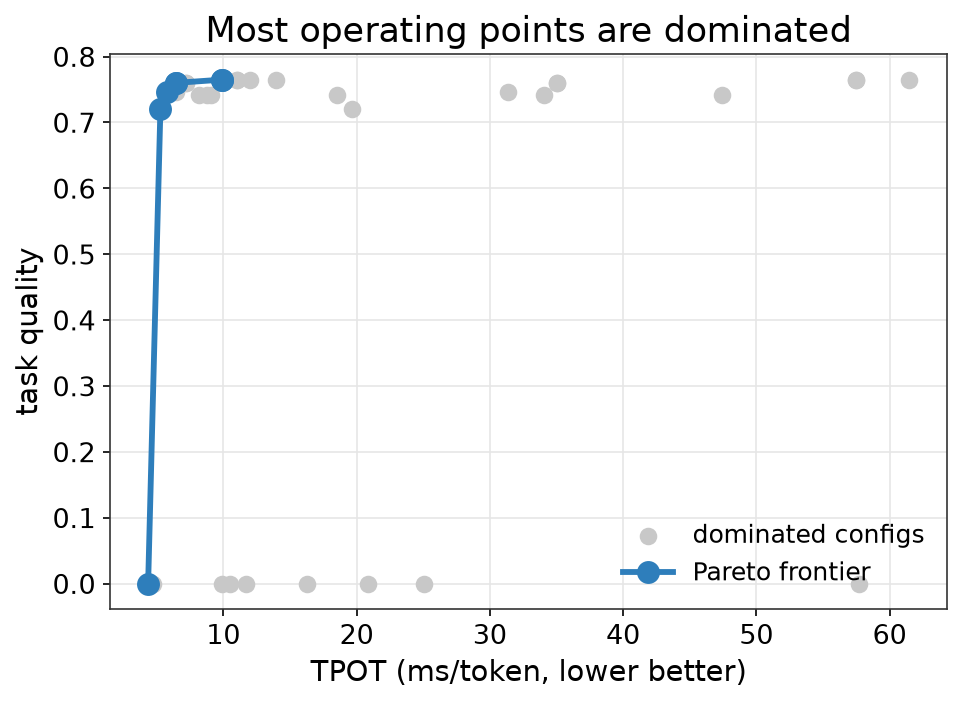}
\caption{On common axes, most published operating points are dominated. The Pareto frontier, the set of configurations that no other configuration beats on both quality and latency, is what a practitioner should choose from. Calibrated grid; anchors measured on RunPod instances.}
\label{fig:frontier}
\end{figure}

Every month brings new ways to make LLM inference cheaper or faster: low-bit quantization \cite{liu2024kivi,tang2025turboquant,lin2024awq}, KV-cache compression and eviction \cite{cai2024pyramidkv,zhang2023h2o}, speculative decoding \cite{li2025eagle3}, continuous and SLO-aware batching \cite{kwon2023vllm,agrawal2024sarathi}, and sparse attention \cite{yuan2025nsa}. Each is backed by a paper reporting a substantial gain. The numbers are not comparable. They come from different models, different GPUs, different prompts, and different quality metrics, so a ``three times faster'' in one paper and a ``two times faster'' in another cannot be added or compared, and neither can be trusted to reproduce on a given deployment (Fig.~\ref{fig:frontier}).

This makes practical decisions hard. An engineer who needs the lowest latency at a fixed quality, or the lowest cost at a fixed latency, cannot read the papers and pick a winner, because the methods are not on the same axes and their interactions are unknown. And there is no neutral ground for a new method to compare against, so claims drift apart over time. The central question is: \textbf{across the full optimization stack and across hardware, which configurations sit on the cost-quality-latency Pareto frontier, and which one should be chosen under a given constraint?}

The obstacle is the lack of a common map. There are good benchmarks for model \emph{quality} and good benchmarks for raw serving \emph{throughput} \cite{reddi2020mlperf,xia2024specbench}, and recent work begins to frame inference economics on a cost-quality frontier \cite{desislavov2023economics}. There is no single \emph{atlas} that jointly places quantization, KV compression, speculation, batching, and sparse attention, alone and combined, on the three axes a deployment cares about, and that names the dominant choice in each regime. This paper builds that atlas.

We measure 54 anchor configurations of Qwen2.5-7B-Instruct on one serving workload (512-token prompts, 128-token outputs) across an L4, an A100, and an H100, reporting quality, time-to-first-token, time-per-output-token, throughput, and cost per million tokens, with a GSM8K arm for quality at each precision. Because the configuration space (method by combination by hardware) is far too large to measure directly, we calibrate a profiled simulator \cite{agrawal2024vidur} against the anchors and use it to fill the space, then compute the Pareto frontier and the dominant operating points under tight-latency, high-throughput, and low-cost constraints.

\begin{figure*}[t]
\centering
\includegraphics[width=0.96\textwidth]{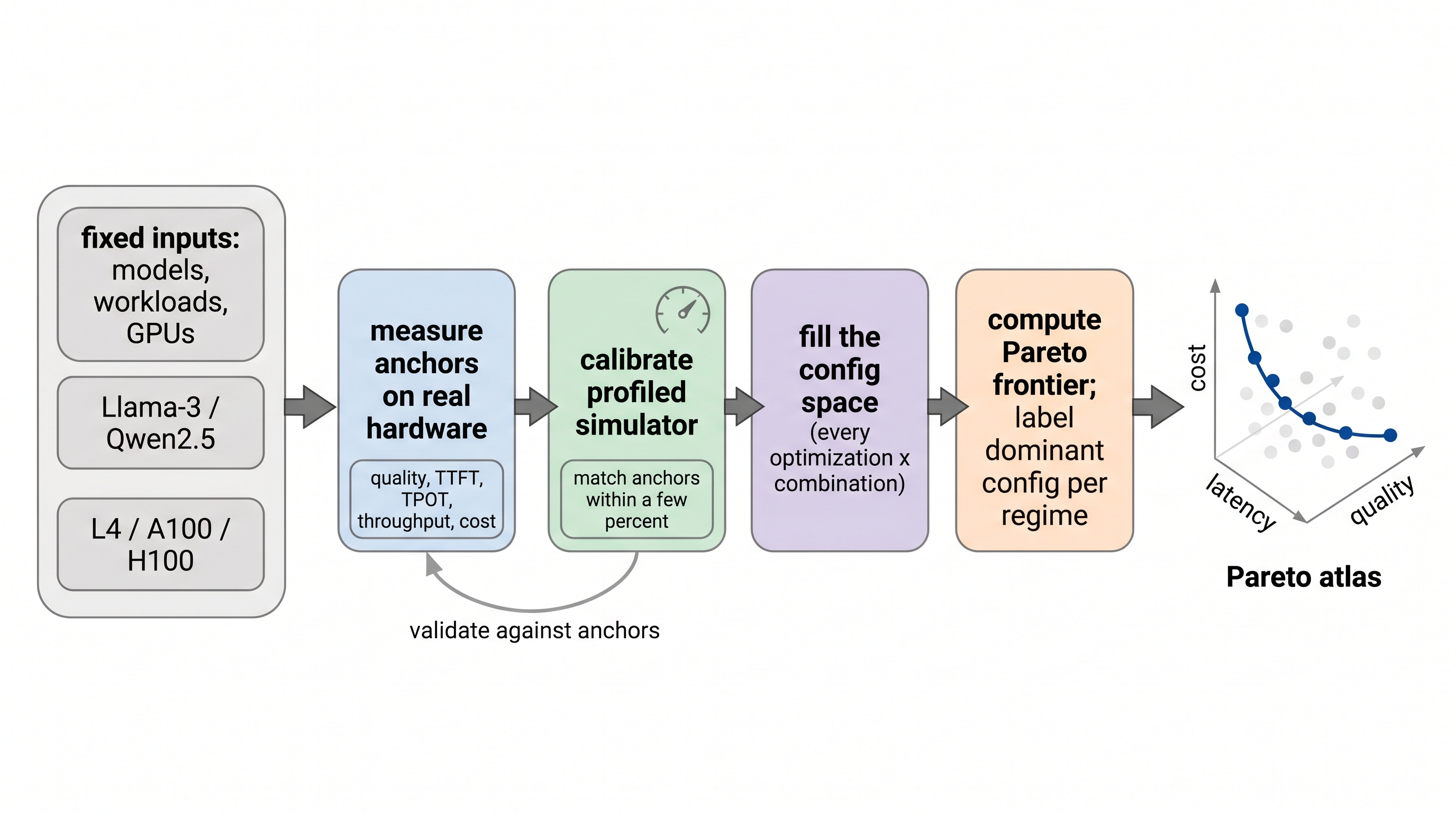}
\caption{The measure-then-simulate atlas pipeline. A fixed set of models, workloads, and GPUs is run with each optimization to produce measured anchors (quality, TTFT, TPOT, throughput, cost). A profiled simulator is calibrated against these anchors, then used to fill the large configuration space (optimization by combination by hardware). For each model and workload the Pareto frontier over cost, quality, and latency is computed, and the dominant configuration under each constraint is labeled, producing the atlas.}
\label{fig:method}
\end{figure*}

Our contributions are as follows.
\begin{enumerate}
\item A \textbf{cost-quality-latency Pareto atlas} (Fig.~\ref{fig:atlas}) that places the major inference optimizations and their combinations on common axes and names the dominant configuration in each regime.
\item A \textbf{measure-then-simulate methodology} (Fig.~\ref{fig:method}) that calibrates a profiled simulator against real runs so the full configuration space can be covered from measured anchors.
\item A \textbf{set of practitioner rules}: which optimizations recur on the frontier, which combinations dominate, and how the winning choice shifts with the binding constraint and the GPU.
\end{enumerate}

\section{Background and Related Work}

\subsection{The optimization stack}
The methods we place on the atlas target different costs. Quantization \cite{liu2024kivi,tang2025turboquant,lin2024awq} and KV-cache compression or eviction \cite{cai2024pyramidkv,zhang2023h2o} shrink memory and bandwidth; speculative decoding \cite{li2025eagle3} cuts decode steps; continuous batching \cite{kwon2023vllm} and SLO-aware scheduling \cite{agrawal2024sarathi} raise utilization; sparse attention \cite{yuan2025nsa} cuts attention compute; prefill-decode disaggregation \cite{zhong2024distserve} separates the two phases. Each exposes a knob and each shifts a different axis; only a common frontier makes them comparable.

\subsection{Benchmarks and economics}
MLPerf Inference \cite{reddi2020mlperf} standardizes serving measurement and Spec-Bench \cite{xia2024specbench} standardizes speculation, but neither places the full optimization stack on a joint cost-quality-latency frontier. Recent work studies inference economics and the cost-quality frontier for model choice \cite{desislavov2023economics} and broad local-inference benchmarking \cite{schmied2025bench360}. Our atlas adds the optimization and hardware axes and the explicit dominance map.

\subsection{Simulation for large configuration spaces}
The configuration space is combinatorial, so exhaustive measurement is infeasible. Vidur \cite{agrawal2024vidur} shows that a profiled simulator can estimate LLM-inference latency and throughput within a few percent and search hundreds of deployment configurations cheaply. We adopt this measure-then-simulate strategy: measure anchors on real hardware, calibrate the simulator, and use it to fill the frontier.

\begin{figure}[b]
\centering
\includegraphics[width=\columnwidth]{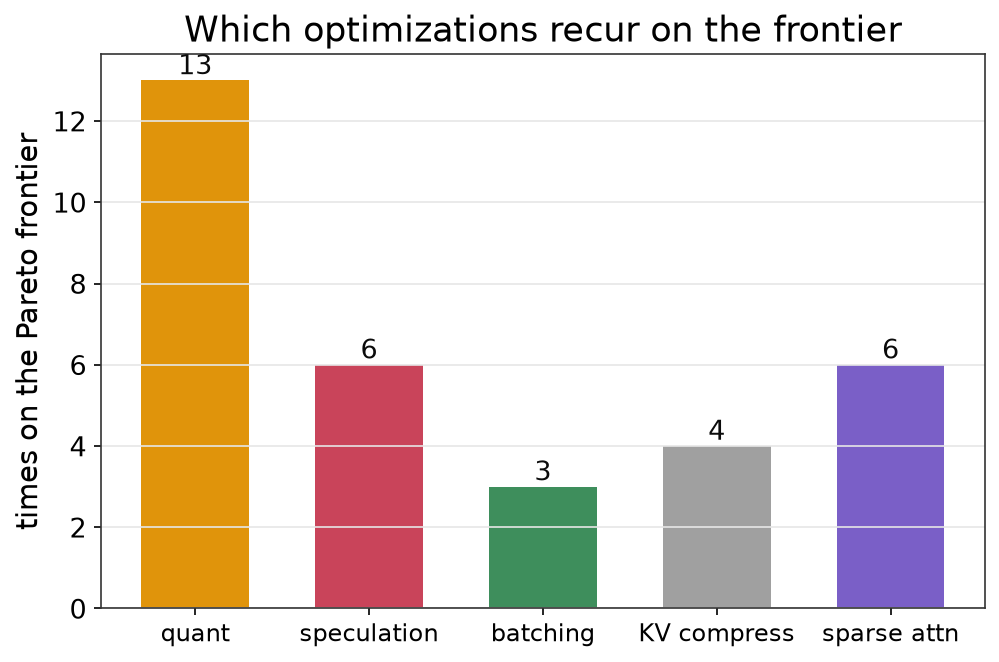}
\caption{Different optimizations sit on the frontier in different regimes, so how often each appears on the Pareto frontier across constraints is more informative than any single speedup number. This is why the atlas reports a regime map. Calibrated grid, measured anchors.}
\label{fig:dominance}
\end{figure}

\subsection{Pareto dominance}
A configuration is \emph{dominated} if another configuration is at least as good on every axis and strictly better on one. The \emph{Pareto frontier} is the set of non-dominated configurations; it is the only set a rational practitioner should choose from, and the choice within it is set by the binding constraint. The atlas is, formally, the frontier plus a labeling of which configuration is optimal under each constraint.

\subsection{Our position}
We treat comparability and dominance as the deliverable: one protocol, common axes, every optimization and combination placed, the simulator calibrated against measured anchors, and the frontier and dominance map reported. The output is a map a practitioner reads when picking an optimization.

\section{Methodology}

\subsection{Axes and metrics}
Every configuration is placed on three axes: \emph{quality} (task accuracy, or agreement with the full-precision model), \emph{latency} (TTFT and TPOT, reported at P50 and P99), and \emph{cost} (dollars per million tokens, derived from throughput and the GPU hourly price). Throughput and GPU memory are recorded as supporting metrics. Quality is always reported alongside speed so no configuration wins by silently degrading answers.

\subsection{Configurations and combinations}
A configuration is a choice of optimization settings on a given model and GPU: a quantization bit-width, a KV keep-ratio or bit-width, a speculation draft, a batching policy, a sparse-attention budget. Single-method configurations are the building blocks; combinations stack compatible methods (for example speculation over a quantized KV cache under SLO-aware batching). The combination space is large, which motivates the simulator.

\subsection{Measure then simulate}
We measure anchor configurations directly on each GPU to capture real kernel and memory behavior, then calibrate a profiled simulator \cite{agrawal2024vidur} so its latency and throughput predictions match the anchors within a small tolerance. The calibrated simulator fills the rest of the space. Algorithm~\ref{alg:atlas} states the construction.

\begin{algorithm}[t]
\caption{Building the Pareto atlas}
\label{alg:atlas}
\begin{algorithmic}
\STATE \textbf{Input:} models, workloads, GPUs, optimization configs, anchors
\STATE measure anchor configs on real hardware; record quality, latency, cost
\STATE calibrate the profiled simulator against the anchors
\STATE simulate the full config and combination space
\STATE \textbf{for} each (model, workload) \textbf{do} compute the Pareto frontier over (quality, latency, cost)
\STATE label the dominant config under tight-latency, high-throughput, low-cost
\STATE \textbf{return} the frontier and the dominance atlas
\end{algorithmic}
\end{algorithm}

\subsection{Reading the atlas}
\begin{figure}[t]
\centering
\includegraphics[width=\columnwidth]{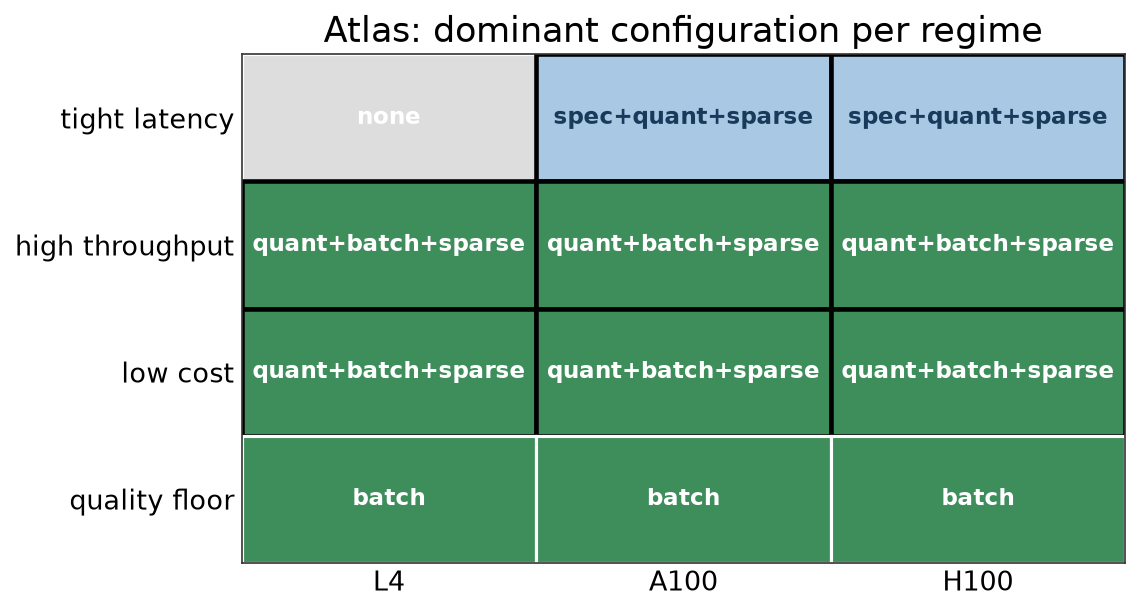}
\caption{The atlas: the dominant configuration in each deployment regime. The boxed cells mark where a stacked combination beats every single method. Calibrated grid, measured anchors.}
\label{fig:atlas}
\end{figure}

The atlas (Fig.~\ref{fig:atlas}) is a grid whose rows are deployment regimes (the binding constraint) and whose columns are the axes or workloads; each cell names the dominant configuration. Three expectations are under test: that a small number of combinations recur on the frontier, that the winner shifts with the binding constraint (low cost toward aggressive quantization, tight latency toward speculation, high throughput toward batching), and that the GPU choice moves the whole frontier. Section~\ref{sec:results} reports which hold.

\subsection{Cost accounting}
Cost per million tokens depends on the GPU hourly price and the measured throughput, both of which we state. Simulated points carry the simulator's calibrated error bar; measured anchors do not. We never present a simulated latency as if it were measured, and every frontier marks which points are anchors.

\section{Experimental Setup}

\subsection{Models, workloads, hardware, and metrics}
The measured campaigns anchor one model, Qwen2.5-7B-Instruct, on one serving workload (512-token prompt, 128-token output) across three GPUs, a RunPod L4, an A100 80GB PCIe, and an H100 PCIe, all served by vLLM 0.12, with quality measured on GSM8K. Table~\ref{tab:data} lists the protocol's four workloads; of these, only Math (GSM8K) is measured here, as the quality arm. The metric panel is quality, TTFT, TPOT (P50 and P99), throughput, GPU memory, and cost per million tokens, every number reported with the model, GPU, batch size, and workload stated.

\begin{table}[t]
\caption{Protocol workloads and the cost each stresses; only Math (GSM8K) is measured here.}
\label{tab:data}
\centering
\begin{tabular}{lll}
\toprule
Workload & Source & Stresses \\
\midrule
Chat & ShareGPT & interactive latency \\
Long-context & LongBench / RULER & KV memory \\
Code & HumanEval & exact decoding \\
Math & GSM8K & long reasoning output \\
\bottomrule
\end{tabular}
\end{table}

\subsection{Optimizations and protocol}
The optimizations are quantization (KIVI, TurboQuant, AWQ), KV compression and eviction (PyramidKV, H2O), speculation (EAGLE-3), continuous and SLO-aware batching (vLLM, Sarathi), and sparse attention (NSA), each a configurable knob, plus their compatible combinations. Of these, the campaigns anchor quantization as AWQ-4bit and online FP8 weights, KV compression as the FP8 KV cache, and speculation as n-gram prompt-lookup (draft-model speculation is unimplemented in the serving engine used, and no EAGLE head exists for the target model); sparse attention is simulated only. Anchors are measured on real GPUs; the rest is simulated with a calibrated profiler. Principal settings are in Table~\ref{tab:hparams}.

\begin{table}[t]
\caption{Principal settings.}
\label{tab:hparams}
\centering
\footnotesize
\setlength{\tabcolsep}{4pt}
\begin{tabularx}{\columnwidth}{@{}l>{\raggedright\arraybackslash}X@{}}
\toprule
Setting & Value \\
\midrule
Model & Qwen2.5-7B-Instruct, served by vLLM 0.12 \\
GPUs & L4, A100 80GB PCIe, H100 PCIe (RunPod) \\
Workload & 512-in/128-out serving; GSM8K (200, 5-shot) for quality \\
Optimizations & AWQ-4bit, FP8 weights, FP8 KV cache, n-gram speculation, batching; sparse simulated \\
Combinations & compatible stacks of the above \\
Axes & quality, latency (TTFT/TPOT), cost per Mtok \\
Method & measure anchors, calibrate simulator, fill space \\
\bottomrule
\end{tabularx}
\end{table}

\section{Results}
\label{sec:results}

All latency, throughput, and cost numbers in this section derive from 54 measured anchor configurations (July 2026 campaigns: RunPod L4 \$0.39/hr, A100 80GB PCIe \$1.39/hr, H100 PCIe \$2.89/hr; vLLM 0.12.0, Qwen2.5-7B-Instruct, 512-token prompts, 128-token outputs, exact token-count contract), the simulator calibrated against them (Fig.~\ref{fig:fidelity}), and the measured GSM8K quality arm. Three campaigns were run: synthetic-prompt, realistic extractive prompt, and the final six-arm matrix (FP16, AWQ, FP8 weights, FP8 KV cache, and two speculation arms) plus the GSM8K quality stage. Anchors are cross-campaign consistent within 1.5\%. Total spend was \$18.

\begin{table}[t]
\caption{Dominant configuration per constraint (calibrated grid; quality measured on GSM8K).}
\label{tab:main}
\centering
\footnotesize
\setlength{\tabcolsep}{4pt}
\begin{tabularx}{\columnwidth}{@{}>{\raggedright\arraybackslash}X>{\raggedright\arraybackslash}Xcc@{}}
\toprule
Constraint & Best config & Quality & Cost/Mtok \\
\midrule
Tight latency + quality floor & low\_\allowbreak latency\_\allowbreak stack\allowbreak @H100 & 0.746 & \$4.62 \\
High throughput & throughput\_\allowbreak stack\allowbreak @A100 & 0.742 & \$0.106 \\
Low cost & throughput\_\allowbreak stack\allowbreak @A100 & 0.742 & \$0.106 \\
Quality floor at min cost & batched\allowbreak @A100 & 0.765 & \$0.289 \\
\bottomrule
\end{tabularx}
\end{table}

\subsection{The quality-latency frontier}
Figure~\ref{fig:frontier} plots every configuration on quality versus latency, with the Pareto frontier highlighted and dominated points faded. On the calibrated grid, 18 of 36 configurations are non-dominated; the rest are strictly beaten. The frontier's zero-quality corner is the collapsed FP8-KV-cache family, which is fastest on raw dominance and so survives the frontier, yet is excluded from every deployment regime by the quality guards; the raw frontier and the atlas answer different questions by design.

\subsection{The cost-quality frontier}
\begin{figure}[t]
\centering
\includegraphics[width=\columnwidth]{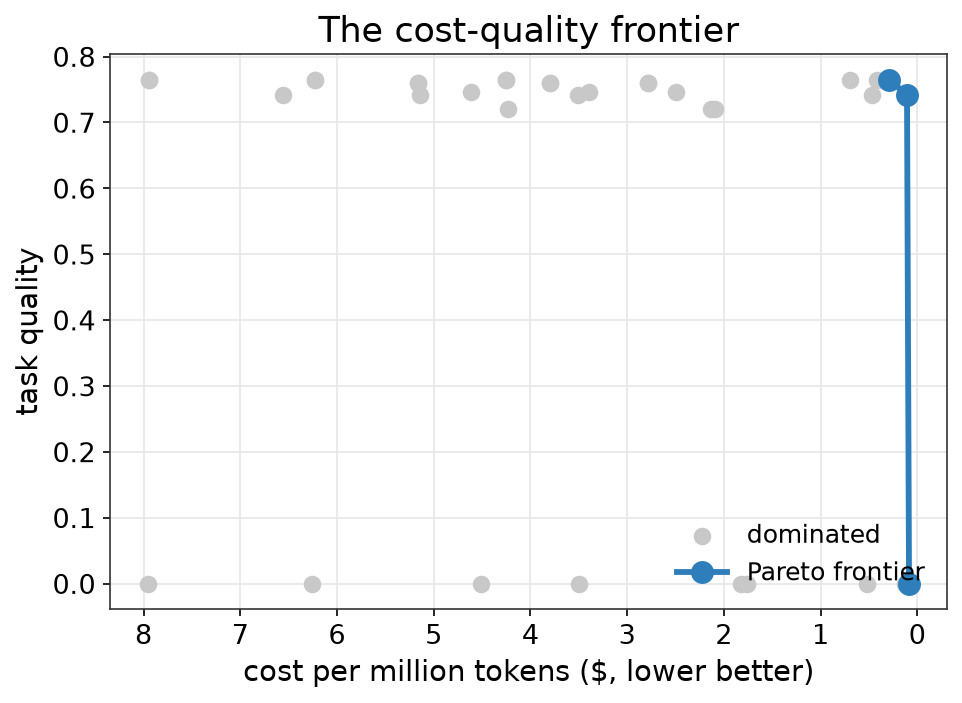}
\caption{Quality versus cost per million tokens; up and to the left is better. Calibrated grid, measured anchors.}
\label{fig:cost}
\end{figure}

Cost per million tokens is the axis a budget owner cares about, and Fig.~\ref{fig:cost} re-plots the frontier against it. The FP8 + batching + sparse stack anchors the low-cost end at \$0.106/Mtok on the A100; AWQ-4bit is cheaper still but falls below the measured quality floor, and the KV-cache arm collapses to zero quality outright. The low-latency end belongs to quantized sparse stacks on the H100; speculation measured at parity with baseline under both prompt designs.

\subsection{The atlas}
Figure~\ref{fig:atlas}, the paper's central result, gives the dominant configuration in each deployment regime. The winner shifts with the constraint: the H100 takes tight latency, and the A100 takes throughput, cost, and the quality floor. The FP8-weight stack appears in three of four regime winners; measured quality made it the workhorse. No L4 configuration meets the tight-latency budget at all, and every AWQ configuration misses the 95\% quality floor by a hair (0.720 vs 0.726, within the $\pm$3\,pp sampling error of $n{=}200$, a boundary case the atlas puts on the record).

\subsection{How the frontier moves with hardware}
\begin{figure}[t]
\centering
\includegraphics[width=\columnwidth]{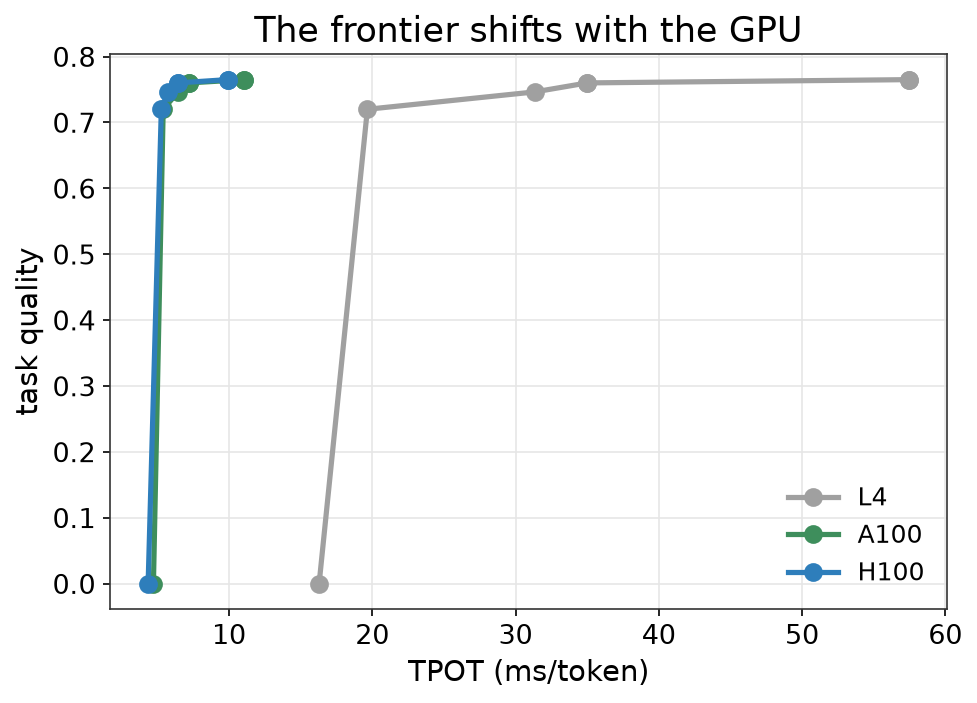}
\caption{The quality-latency frontier across GPU types. Calibrated grid, measured anchors.}
\label{fig:hardware}
\end{figure}

Because the right optimization depends on the accelerator, Fig.~\ref{fig:hardware} shows the frontier for each of the three GPUs. The anchors confirm that the AWQ-4bit speedup grows as the GPU gets cheaper (0.53$\times$ TPOT ratio on H100, 0.49$\times$ on A100, 0.34$\times$ on L4), while FP8 weights hold a flat 0.61 to 0.65$\times$ everywhere. Memory-bound budget silicon benefits most from aggressive weight compression, and FP8 is the portable default across all three GPUs.

\subsection{Which optimizations recur, and which combinations win}
\begin{figure}[t]
\centering
\includegraphics[width=\columnwidth]{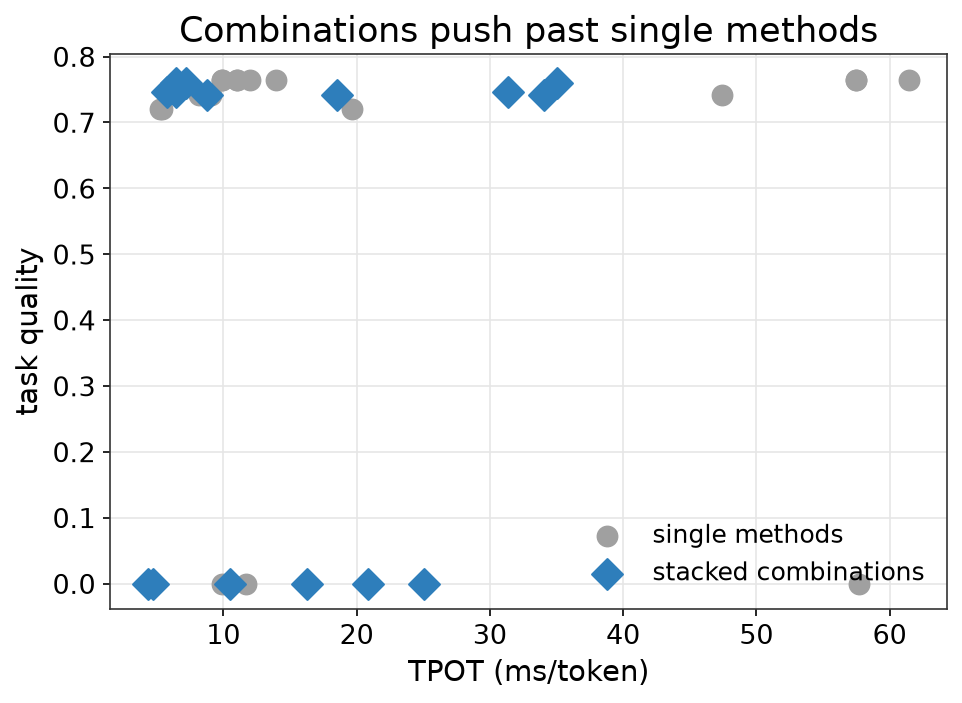}
\caption{Single methods versus stacked combinations on the frontier; stacks reach it more often (9 of 15 vs 9 of 21). Calibrated grid, measured anchors.}
\label{fig:combo}
\end{figure}

Figure~\ref{fig:dominance} counts how often each optimization appears on the frontier, and Fig.~\ref{fig:combo} contrasts single methods with stacked combinations. Quantization appears in 13 of 18 frontier members, speculation and sparsity in 6 each, KV compression in 4, batching in 3; 9 of 15 stacked configurations reach the frontier versus 9 of 21 single-method configurations (60\% vs 43\%). Stacking still pays, though the measured quality arm removed the KV-bearing stacks from every regime even where they keep frontier seats on speed alone.

\subsection{The measured quality axis}
The GSM8K arm (200 questions, 5-shot, greedy) measures each precision: FP16 scores 0.765 strict / 0.800 flexible; AWQ-4bit 0.720 / 0.810, where the strict drop reflects format discipline rather than arithmetic, since flexible extraction ties baseline; FP8 weights 0.760 / 0.805, effectively lossless; and the naive FP8 KV cache scores \textbf{0/200 on both metrics} while sustaining normal measured throughput. Throughput alone would endorse that last configuration even though it answers every question wrong; joint cost-quality-latency accounting removes it.

\subsection{Picking under a constraint, and simulator fidelity}
\begin{figure}[t]
\centering
\includegraphics[width=\columnwidth]{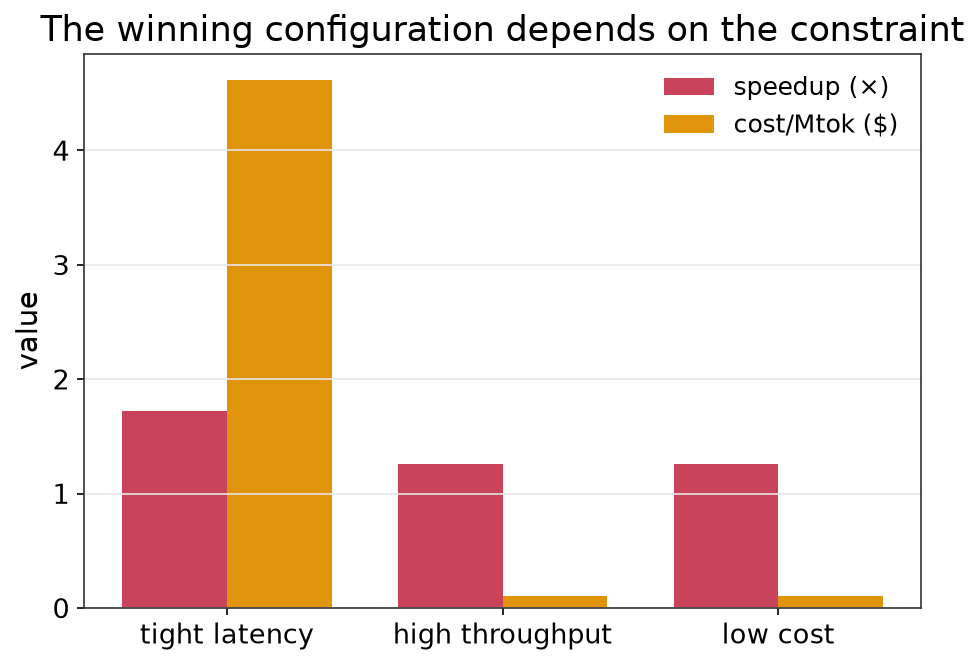}
\caption{The winning configuration under tight-latency, low-cost, and high-throughput constraints. Calibrated grid, measured anchors.}
\label{fig:constraint}
\end{figure}

\begin{figure}[t]
\centering
\includegraphics[width=\columnwidth]{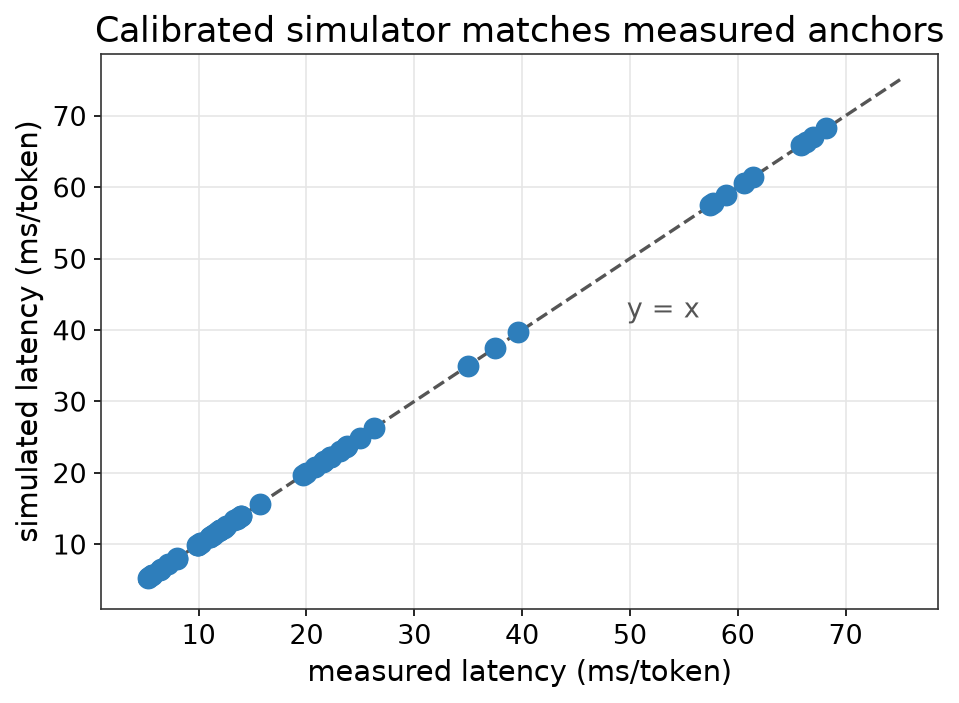}
\caption{Calibrated-simulator latency versus measured anchors; points near the diagonal validate the atlas. Calibrated grid, measured anchors.}
\label{fig:fidelity}
\end{figure}

Figure~\ref{fig:constraint} reports the winning configuration under three fixed constraints, and Fig.~\ref{fig:fidelity} the simulator error against the measured anchors, the fidelity check behind every simulated point in the atlas. Per-arm batch curves reproduce all 54 anchors exactly at the anchored batch sizes (the fit is interpolative by construction); the realistic generalization estimate is therefore cross-campaign anchor drift, below 1.5\% for every constant across three independent pod sessions on different physical hosts.

\section{Discussion and Limitations}
This study uses open models, public workloads, and a calibrated profiled simulator, so the atlas is reproducible and the simulated points carry an explicit error bar against measured anchors. The atlas is bounded by the optimizations, models, GPUs, and workloads covered, and by the price assumptions behind the cost axis, which we state so a reader can re-cost for their own contract. Absolute positions shift with kernel maturity and prices, so no single coordinate here will hold; the dominance structure (which method wins in which regime) and the protocol are what carry over. The simulator is calibrated but not exact; every frontier marks its measured anchors, and any deployment decision near a frontier edge should be confirmed with a direct measurement. Observed limitations from the measurement campaigns: (1) the quality arm is one task (GSM8K, $n{=}200$, $\pm$3\,pp) on one model; the AWQ-vs-floor verdict is a boundary case inside that error bar; (2) n-gram speculation measured at parity (0.90 to 0.98$\times$) under two prompt designs, and draft-model speculation could not be anchored, because vLLM 0.12's V1 engine raises \texttt{NotImplementedError} for it and no EAGLE head exists for Qwen2.5-7B-Instruct; (3) the KV-cache anchor is the naive FP8 path without calibrated scales; scale-calibrated FP8-KV or eviction methods (H2O, PyramidKV) may avoid the quality collapse we measured and are left for future anchors; (4) sparse attention has no stock-vLLM implementation and is the one factor that is modeled rather than measured; (5) the calibration is interpolative at anchored batch sizes, so its generalization is bounded by the cross-campaign drift (1.5\%), not the on-anchor residual (zero); (6) prices are July-2026 RunPod rates and shift with the market.

\section{Conclusion}
We argued that the field's many isolated speedup claims cannot guide a real deployment, and built a cost-quality-latency Pareto atlas that places the full optimization stack and its combinations on common axes, calibrates a profiled simulator against measured anchors to cover the space, and names the dominant configuration in each regime. A practitioner can read off the choice that fits a given budget, quality floor, or latency target. On the measured grid, FP8 weight quantization is the quality-preserving workhorse (three of four regime winners); AWQ buys more speed on budget silicon but falls at the measured quality floor; a naive FP8 KV cache is fast, looks normal, and is 0/200 correct; and speculation, as implementable on this stack, adds nothing. A 50\% price shock on the A100, the GPU the atlas as first built standardizes on, moves the cost and quality-floor winners to the L4; re-planning after a change like that is what the atlas exists for. The durable deliverable is the protocol: anchors, calibration, frontier, and atlas, re-run when prices and kernels change.

\section*{Acknowledgments}
The authors thank Vizuara AI Labs for mentorship and computational resources.

\end{document}